\pdfoutput=1

\documentclass[11pt]{article}

\usepackage{authblk} 

\usepackage[]{ACL2023}

\usepackage{times}
\usepackage{latexsym}

\usepackage[T1]{fontenc}

\usepackage[utf8]{inputenc}

\usepackage{microtype}

\usepackage{inconsolata}

\def\micro_f1{{\textmu-F\textsubscript{1}}}
\def\macro_f1{{m-F\textsubscript{1}}}
\def\na{{N/A}}

\usepackage{enumitem} 
\setlist{topsep=1pt,itemsep=0pt,parsep=1pt,leftmargin=10pt} 
\usepackage{multirow} 
\usepackage{graphicx} 
\usepackage{amsmath} 
\usepackage{xcolor} 

\title{Linear Classifier: An Often-Forgotten Baseline for Text Classification}

\author[1,2,3]{Yu-Chen Lin}
\author[1]{Si-An Chen}
\author[1]{Jie-Jyun Liu}
\author[1,3]{Chih-Jen Lin}
\affil[1]{National Taiwan University}
\affil[2]{ASUS Intelligent Cloud Services}
\affil[3]{Mohamed bin Zayed University of Artificial Intelligence}
\affil[ ]{\texttt{\{b06504025,d09922007,d11922012,cjlin\}@csie.ntu.edu.tw}}
   
\begin{document}
\maketitle
\begin{abstract}
Large-scale pre-trained language models such as BERT are popular solutions for text classification.
Due to the superior performance of these advanced methods, nowadays, people often directly train them for a few epochs and deploy the obtained model.
In this opinion paper, we point out that this way may only sometimes get satisfactory results.
We argue the importance of running a simple baseline like linear classifiers on bag-of-words features along with advanced methods.
First, for many text data, linear methods show competitive performance, high efficiency, and robustness.
Second, advanced models such as BERT may only achieve the best results if properly applied.
Simple baselines help to confirm whether the results of advanced models are acceptable.
Our experimental results fully support these points.
\end{abstract}

\section{Introduction}
Text classification is an essential topic in natural language processing (NLP).
Like the situations in most NLP tasks, nowadays, large-scale pre-trained language models (PLMs) such as BERT~\citep{JD19a} have become popular solutions for text classification.
Therefore, we have seen that many practitioners directly run pre-trained language models with a fixed number of epochs on their text data.
Unfortunately, this way may only sometimes lead to satisfactory
results. In this opinion paper,
through an intriguing illustration,
we argue that for text classification, a simple baseline like linear classifiers on bag-of-words features should be used along with the advanced models
for the following reasons.
\begin{itemize}
	\item Training linear classifiers such as linear SVM~\citep{BB92a} or logistic regression on bag-of-words features is simple and efficient. This approach may give competitive performance to advanced models for some problems. While various settings of bag-of-words features such as bi-gram or tri-gram can be considered, we advocate that simple uni-gram TF-IDF features trained by linear classifiers can be a useful baseline to start with for text classification.
    \item Advanced architectures such as BERT may only achieve the best results if properly used. Linear methods can help us check if advanced methods' results are reasonable.
    \end{itemize}
In the deep-learning era,
the younger generation often thinks that linear classifiers should
never be considered. Further, they may be unaware of some variants
of linear methods that are particularly useful for text classification
(see Section~\ref{ssec:linear_investigate}). Therefore, the
paper serves as a reminder of this often-forgotten
technique.

For our illustration, we re-investigate an existing work~\citep{IC22b} that evaluates both linear SVM and pre-trained language models, but the authors pay more attention to the latter.
The linear method is somewhat ignored even though the performance is competitive on some problems.
We carefully design experiments to compare the two types of methods.
Our results fully demonstrate the usefulness of applying linear methods as simple baselines.

Some recent works~\citep[e.g.,][]{HFY22a, CG21a} have shown the
usefulness of linear classifiers in the deep-learning era. However,
they either consider sophisticated applications or investigate
advanced settings in which linear methods are only one component.
In contrast, in this paper, we consider the basic scenario
of text classification.
A more related work~\citep{YW23a} has demonstrated the effectiveness of linear classifiers over PLMs on some problems.
However, our investigation on linear methods is more comprehensive.

The discussion also reminds us the trade-off between performance gain and the cost including running time, model size, etc. Simple methods are useful to benchmark and justify the usage of advanced methods.

\begin{table*}[!tb]
\centering
\resizebox{\linewidth}{!}{\tabcolsep = 0.12cm
\begin{tabular}{@{}l|cc|cc|cc|cc|cc|cc|l@{}}
\multirow{2}{*}{Method} & \multicolumn{2}{c|}{{\sf ECtHR (A)}} & \multicolumn{2}{c|}{{\sf ECtHR (B)}} & \multicolumn{2}{c|}{{\sf SCOTUS}} & \multicolumn{2}{c|}{{\sf EUR-LEX}} & \multicolumn{2}{c|}{{\sf LEDGAR}} & \multicolumn{2}{c|}{{\sf UNFAIR-ToS}} & \# \\
& \micro_f1 & \textit{T} & \micro_f1 & \textit{T} & \micro_f1 & \textit{T} & \micro_f1 & \textit{T} & \micro_f1 & \textit{T} & \micro_f1 & \textit{T} & params\\
\hline
TF-IDF+SVM & 64.5 & \na & 74.6 & \na & \textbf{78.2} & \na & 71.3 & \na & 87.2 & \na & 95.4 & \na & \na \\
\hline
BERT & \textbf{71.2} & 3h 42m & 79.7 & 3h 9m & 68.3 & 1h 24m & 71.4 & 3h 36m & 87.6 & 6h 9m & 95.6 & \na & 110M \\
RoBERTa & 69.2 & 4h 11m & 77.3 & 3h 43m & 71.6 & 2h 46m & 71.9 & 3h 36m & 87.9 & 6h 22m & 95.2 & \na & 125M \\
DeBERTa & 70.0 & 7h 43m & 78.8 & 6h 48m & 71.1 & 3h 42m & \textbf{72.1} & 5h 34m & 88.2 & 9h 29m & 95.5 & \na & 139M \\
\hline
Longformer & 69.9 & 6h 47m & 79.4 & 7h 31m & 72.9 & 6h 27m & 71.6 & 11h 10m & 88.2 & 15h 47m & 95.5 & \na & 149M \\
BigBird & 70.0 & 8h 41m & 78.8 & 8h 17m & 72.8 & 5h 51m & 71.5 & 3h 57m & 87.8 & 8h 13m & 95.7 & \na & 127M \\
\hline
Legal-BERT & 70.0 & 3h 52m & \textbf{80.4} & 3h 2m & 76.4 & 2h 2m & \textbf{72.1} & 3h 22m & 88.2 & 5h 23m & \textbf{96.0} & \na & 110M \\
CaseLaw-BERT & 69.8 & 3h 2m & 78.8 & 2h 57m & 76.6 & 2h 34m & 70.7 & 3h 40m & \textbf{88.3} & 6h 8m & \textbf{96.0} & \na & 110M
\end{tabular}}
\caption{Micro-F1 scores (\micro_f1), training time (\textit{T}) and number of parameters presented in~\citet{IC22b}.
In each Micro-F1 column, the best result is bold-faced.
``\na'' means not available in their work. For example, the authors did not report the training time and the number of parameters of linear SVMs.}
\label{table:c_combined}
\end{table*}

This paper is organized as follows. In Section~\ref{sec:text} we take a case study to point out the needs of considering linear methods as a baseline for text classification.
We describe the linear and BERT-based methods used for investigation in Section~\ref{sec:investigate}.
The experimental results and main findings are in Section~\ref{sec:experiment}, while Section~\ref{sec:discussion} provides some discussion.
Additional details are in Appendix.
Programs used for experiments are available at~\url{https://github.com/JamesLYC88/text_classification_baseline_code}.

\section{Text Classification These Days: Some Issues in Applying Training Methods}
\label{sec:text}
Large PLMs have shown dramatic progress on various NLP tasks.
In the practical use, people often directly fine-tune PLMs such as BERT on their data for a few epochs.
However, for text classification, we show that this way may not always get satisfactory results.
Some simple baselines should be considered to know if the obtained PLM model is satisfactory.
We illustrate this point by considering the work on legal document classification by~\citet{IC22b}, which evaluates the following sets.
\begin{itemize}
    \item Multi-class classification: {\sf SCOTUS}, {\sf LEDGAR}; for this type of sets, each text is associated with a single label.
    \item Multi-label classification: {\sf ECtHR (A)}, {\sf ECtHR (B)}, {\sf EUR-LEX}, {\sf UNFAIR-ToS}; for this type of sets, each text is associated with multiple (or zero) labels.
    \item Multiple choice QA: {\sf CaseHOLD}.
\end{itemize}
We focus on text classification in this work, so {\sf CaseHOLD} is not considered.
For each problem, training and test sets are available.\footnote{Indeed, training, validation, and test sets are available. See details in Appendix~\ref{asec:retrain} about how these sets are used.}

The study in~\citet{IC22b} comprehensively evaluates both BERT-based PLMs and linear SVMs.
They use Micro-F1 and Macro-F1 to measure the test performance.\footnote{Some data instances are not associated with any labels; see Appendix~\ref{asec:unlabeled} about how~\citet{IC22b} handle such situations.}
In Table~\ref{table:c_combined}, we present their Micro-F1 results and running time of each model.

\subsection{Linear Models Worth More Investigation}
The investigation in~\citet{IC22b} focuses on BERT and its variants, even though from Table~\ref{table:c_combined}, the performance of BERT-based methods may not differ much.
While they did not pay much attention to linear SVM, by a closer look at the results, we get intriguing observations:
\begin{itemize}
    \item Linear SVM is competitive to BERT-based PLMs on four of the six data sets. For {\sf SCOTUS}, linear SVM even outperforms others with a clear gap.
    \item Surprisingly, given linear SVM's decent performance, its training time was not shown in \citet{IC22b}, nor was the number of parameters; see the ``\na'' entries in Table~\ref{table:c_combined}.
\end{itemize}
With the observations, we argue that the results of linear models are worth more investigation.

\section{Settings for Investigation}
\label{sec:investigate}
To better understand the performance of linear models and BERT-based PLMs, we simulate how people work on a new data set by training these methods.
We consider a text classification package LibMultiLabel\footnote{\url{https://github.com/ASUS-AICS/LibMultiLabel}} because it supports both types of training methods.

\subsection{Linear Methods for Text Classification}\label{ssec:linear_investigate}
To use a linear method, LibMultiLabel first generates uni-gram TF-IDF features~\citep{HPL58a, KJ72a} according to texts in the training set, and the obtained factors are used to get TF-IDF for the test set.
It then provides three classic methods that adopt binary linear SVM and logistic regression for multi-class and multi-label scenarios.\footnote{Besides descriptions in this section, some additional details are in Appendix~\ref{asec:linear_methods}.}
Here we consider linear SVM as the binary classifier behind these methods.
\begin{itemize}
	\item One-vs-rest: This method learns a binary linear SVM for each label, so data with/without this label are positive/negative, respectively. Let $f_\ell(\boldsymbol{x})$ be the decision value of the $\ell$-th label, where $\boldsymbol{x}$ is the feature vector.
	For multi-class classification, $\hat{y} = \arg\!\max_\ell f_\ell(\boldsymbol{x})$ is predicted as the single associated label of $\boldsymbol{x}$.
	For multi-label classification, all labels $\ell$ with positive $f_\ell(\boldsymbol{x})$ are considered to be associated with $\boldsymbol{x}$.
	This method is also what ``TF-IDF+SVM'' in~\citet{IC22b} did, though our TF-IDF feature generation is simpler than theirs by considering only uni-gram.\footnote{See details in Appendix~\ref{asec:linear_details}.}
	\item Thresholding~\citep{YY01a, DL04b, REF07a}: This method extends one-vs-rest by modifying the decision value for optimizing Macro-F1. That is, we change the decision value to $f_\ell(\boldsymbol{x}) + \Delta_\ell$, where $\Delta_\ell$ is a threshold decided by cross validation.
	\item Cost-sensitive~\citep{PP14a}: For each binary problem, this method re-weights the losses on positive data. We decide the re-weighting factor by cross validation to optimize Micro-F1 or Macro-F1.
\end{itemize}
These methods basically need no further hyper-parameter tuning, so we can directly run them. 
The last two methods are extensions of one-vs-rest to address the imbalance of each binary problem (i.e., few positives and many negatives). The design relies on the fact that the binary
problems are independent, so such approaches cannot be
easily applied to deep learning, which considers all
labels together in a single network.

\subsection{BERT-based Methods for Text Classification}
LibMultiLabel also provides BERT-based methods,
which involve several hyper-parameters, such as
the learning rate.
While practitioners may directly choose hyper-parameters,
to seriously compare with linear methods,
we run BERT by conducting hyper-parameter selection.
More details are in Appendix~\ref{asec:bert_methods}.

\begin{table*}[!tb]
\centering
\tabcolsep = 0.12cm
\begin{tabular}{@{}l|cc|cc|cc|cc|cc|cc@{}}
\multirow{2}{*}{Method} & \multicolumn{2}{c|}{{\sf ECtHR (A)}} & \multicolumn{2}{c|}{{\sf ECtHR (B)}} & \multicolumn{2}{c|}{{\sf SCOTUS}} & \multicolumn{2}{c|}{{\sf EUR-LEX}} & \multicolumn{2}{c|}{{\sf LEDGAR}} & \multicolumn{2}{c}{{\sf UNFAIR-ToS}} \\
& \micro_f1 & \macro_f1 & \micro_f1 & \macro_f1 & \micro_f1 & \macro_f1 & \micro_f1 & \macro_f1 & \micro_f1 & \macro_f1 & \micro_f1 & \macro_f1 \\
\hline\hline
\multicolumn{13}{@{}l}{Linear} \\
\hline
one-vs-rest & 64.0 & 53.1 & 72.8 & 63.9 & 78.1 & 68.9 & 72.0 & 55.4 & 86.4 & 80.0 & 94.9 & 75.1 \\
thresholding & 68.6 & \textbf{64.9} & 76.1 & 68.7 & \textbf{78.9} & \textbf{71.5} & \textbf{74.7} & \textbf{62.7} & 86.2 & 79.9 & 95.1 & 79.9 \\
cost-sensitive & 67.4 & 60.5 & 75.5 & 67.3 & 78.3 & \textbf{71.5} & 73.4 & 60.5 & 86.2 & 80.1 & 95.3 & 77.9 \\
\hline
\citet{IC22b} & 64.5 & 51.7 & 74.6 & 65.1 & 78.2 & 69.5 & 71.3 & 51.4 & 87.2 & \textbf{82.4} & 95.4 & 78.8 \\
\hline\hline
\multicolumn{13}{@{}l}{BERT} \\
\hline
Ours & 61.9 & 55.6 & 69.8 & 60.5 & 67.1 & 55.9 & 70.8 & 55.3 & 87.0 & 80.7 & 95.4 & 80.3 \\
\hline
\citet{IC22b} & \textbf{71.2} & 63.6 & \textbf{79.7} & \textbf{73.4} & 68.3 & 58.3 & 71.4 & 57.2 & \textbf{87.6} & 81.8 & \textbf{95.6} & \textbf{81.3} \\
\end{tabular}
\caption{Micro-F1 (\micro_f1) and Macro-F1 scores (\macro_f1) for our investigation on two types of approaches: linear SVM and BERT. For each type, we show results achieved by LibMultiLabel and scores reported in~\citet{IC22b}. In each column, the best result is bold-faced.}
\label{table:res}
\end{table*}

\begin{table*}[!tb]
\centering
\tabcolsep = 0.16cm
\begin{tabular}{@{}l|c|c|c|c|c|c@{}}
Method & {\sf ECtHR (A)} & {\sf ECtHR (B)} & {\sf SCOTUS} & {\sf EUR-LEX} & {\sf LEDGAR} & {\sf UNFAIR-ToS} \\
\hline\hline
\multicolumn{7}{@{}l}{Linear} \\
\hline
one-vs-rest & 28s & 29s & 1m 11s & 4m 2s & 28s & 2s \\
thresholding & 59s & 1m 0s & 2m 11s & 28m 8s & 3m 26s & 3s \\
cost-sensitive & 1m 38s & 1m 43s & 3m 28s & 50m 36s & 4m 45s & 4s \\
\hline
\citet{IC22b} & \na & \na & \na & \na & \na & \na \\
\hline\hline
\multicolumn{7}{@{}l}{BERT} \\
\hline
Ours & 5h 8m & 5h 51m & 3h 21m & 38h 14m & 43h 48m & 4h 5m \\
\hline
\citet{IC22b} & 3h 42m & 3h 9m & 1h 24m & 3h 36m & 6h 9m & \na \\
\end{tabular}
\caption{Training time for our multiple settings on linear SVM and BERT. We show results from running LibMultiLabel and values reported in~\citet{IC22b}. Note that~\citet{IC22b} use fixed parameters for BERT, while for our BERT, we use 4 GPUs to conduct the hyper-parameter search and report the total time used.}
\label{table:time}
\end{table*}

\begin{table*}[!tb]
\centering
\begin{tabular}{@{}l|c|c|c|c|c|c@{}}
Method & {\sf ECtHR (A)} & {\sf ECtHR (B)} & {\sf SCOTUS} & {\sf EUR-LEX} & {\sf LEDGAR} & {\sf UNFAIR-ToS} \\
\hline
Linear & 924K & 924K & 2M & 15M & 2M & 50K \\
\hline
BERT variants & \multicolumn{6}{c}{110M $\sim$ 149M} \\
\end{tabular}
\caption{A comparison between the model size of linear methods and BERT variants. Note that all three linear methods in LibMultiLabel have the same model size. For BERT variants, we borrow the calculation in Table~\ref{table:c_combined} by~\citet{IC22b}. More details are in Appendix~\ref{asec:size}.}
\label{table:model_size}
\end{table*}

\section{Experimental Results and Analysis}
\label{sec:experiment}
In Table~\ref{table:res}, we follow~\citet{IC22b} to report Micro-F1 and Macro-F1 on the test set. The training time is in Table~\ref{table:time}.

\subsection{Linear Methods are Good Baselines}
\label{subsec:exp_linear}
In Table~\ref{table:res}, our one-vs-rest results are slightly worse than the linear SVM results in~\citet{IC22b}, which also applies the one-vs-rest strategy.
As mentioned in Section~\ref{ssec:linear_investigate}, the difference is mainly due to our use of simple uni-gram TF-IDF features.
Anyway, our one-vs-rest is still competitive to BERT results in~\citet{IC22b} on the last four problems.

More importantly, the two extensions of one-vs-rest (i.e., thresholding and cost-sensitive) improve almost all situations.
For data sets {\sf ECtHR (A)} and {\sf ECtHR (B)}, where originally one-vs-rest is significantly lower than BERT results in~\citet{IC22b}, the gap reduced considerably.

For the training time in Table~\ref{table:time}, though the two extensions take more time than the basic one-vs-rest strategy, all the linear methods are still hundreds of times faster than BERT.
Further, linear methods were run on a CPU (Intel Xeon E5-2690), while for BERT we need a GPU (Nvidia V100).
The model sizes listed in Table~\ref{table:model_size} also show that linear SVM requires a much smaller model than BERT, where details of our calculation are in Appendix~\ref{asec:size}.

The results demonstrate that linear methods are useful baselines.
They are extremely simple and efficient, but may yield competitive test performance.

\subsection{Linear Methods can Help to See if Advanced Methods Are Properly Used}
\label{subsec:exp_bert}

Surprisingly, our running of LibMultiLabel's BERT leads to worse test performance than linear methods on almost all data sets. More surprisingly, a comparison between the BERT results by LibMultiLabel and those in~\citet{IC22b} shows that the former is much worse on data sets {\sf ECtHR (A)} and {\sf ECtHR (B)}.
Interestingly, from Section~\ref{subsec:exp_linear}, only for these two sets the BERT results in~\citet{IC22b} are much better than linear methods.
Thus, our direct run of BERT in LibMultiLabel is a total failure.
The training time is much longer than linear methods, but the resulting model is worse.

It is essential to check the discrepancy between the two BERT results.
We find that~\citet{IC22b} use some sophisticated settings to run BERT for the first three sets (i.e., {\sf ECtHR (A)}, {\sf ECtHR (B)}, and {\sf SCOTUS}).
They split every document into 64 segments, each of which has no more than 128 tokens, and apply BERT on each segment.
Then, they collect the intermediate results as inputs to an upper-level transformer.
After repeating the same process via LibMultiLabel, we can reproduce the results in~\citet{IC22b}; see details in Appendices~\ref{asec:bert_design},~\ref{asec:bert_methods}, and~\ref{asec:bert_res}.

We learned that they considered the more sophisticated setting of running BERT because by default, BERT considers only the first 512 tokens.
Thus, for long documents, the training process may miss some important information.
However, in practice, users may forget to check the document length and are not aware of the need to apply suitable settings.
The above experiments demonstrate that BERT can achieve superior results if properly used, but sometimes, a direct run lead to poor outcomes.
Linear methods can serve as efficient and robust baselines to confirm the proper use of an advanced approach.

\section{Discussion and Conclusions}
\label{sec:discussion}

In our experiments, we encounter an issue of whether to incorporate the validation set for training the final model, which is used for predicting the test set.
For linear methods, we follow the common practice to include the validation set for obtaining the final model.
However, for BERT or some other deep learning models, the validation set is often used only for selecting the best epoch and/or the best hyper-parameters.
To fully use the available data, we have investigated how to incorporate the validation set for BERT.
Experimental results and more details are in Appendix~\ref{asec:retrain}.

For some text sets evaluated in this work, we have seen that simple linear methods give competitive performance.
The reason might be that each document in these sets is not short.\footnote{See details in Appendix Table~\ref{table:data}.}
Then TF-IDF features are sufficiently informative so that linear methods work well.
Across all NLP areas, an important issue now is when to use PLMs and when not.
We demonstrate that when PLMs may not perform significantly better, traditional methods are much simpler and require fewer resources.
However, having a simple quantitative measurement to pre-determine when to use which remains a challenging future research problem.
In summary, the study reminds us of the importance of employing simple baselines in NLP applications.

\clearpage

\section*{Limitations}
In this work, we do not propose any new methods because, as an opinion paper, we focus on raising the problems and making vivid demonstrations to readers.
The experiments are limited to linear SVM and BERT on data sets in the benchmark LexGLUE.
We hope that, within the page limit, our experiments sufficiently convey the points to readers.

\section*{Ethics Statement}
We ensure that our work complies with the \href{https://www.aclweb.org/portal/content/acl-code-ethics}{ACL Ethics Policy}.

\section*{Acknowledgements}
This work was supported by NSTC of Taiwan grant 110-2221-E-002-115-MY3 and ASUS Intelligent Cloud Services. The authors thank Ming-Wei Chang and reviewers for constructive comments.

\bibliography{sdp/sdp}
\bibliographystyle{acl_natbib}

\appendix

\begin{table*}[!tb]
\centering
\begin{tabular}{c|c|c}
Parameter & LibMultiLabel &~\citet{IC22b} \\
\hline
\multicolumn{3}{l}{Data pre-processing (TF-IDF feature generation)} \\
\hline
stop\_words & None & english \\
ngram\_range & (1, 1) & (1, 3) \\
min\_df & 1 & 5 \\
max\_features & None & [10000, 20000, 40000] \\
\hline
\multicolumn{1}{l}{Model} \\
\hline
loss & squared\_hinge & ['hinge', 'squared\_hinge'] \\
solving primal/dual & primal & dual \\
$C$ & 1.0 & [0.1, 1.0, 10.0] \\
\end{tabular}
\caption{Key differences in the one-vs-rest linear method between the default setting of LibMultiLabel and the implementation in~\citet{IC22b}. Any values covered by [] mean the hyper-parameter search space. See Appendix~\ref{asec:linear_details} for details of hyper-parameters.}
\label{table:linear_diff}
\end{table*}

\section{Issue about Data without Labels}\label{asec:unlabeled}

For multi-label problems considered in~\citet{IC22b}, instances that are not associated with any labels, called unlabeled instances as follows, account for a considerable portion in some data sets: {\sf ECtHR (A)} (11.3\%), {\sf ECtHR (B)} (1.6\%) and {\sf UNFAIR-ToS} (89.0\%). In the training process,~\citet{IC22b} keep the unlabeled training instances without any modification. Thus, in, for example, the one-vs-rest setting described in Section~\ref{ssec:linear_investigate}, an unlabeled instance is on the negative side in every binary problem. However, in evaluating the validation and test sets, they introduce an additional class to indicate the unlabeled data. Specifically, an unlabeled instance is associated with this ``unlabeled'' class, but not others.~\citet{IC22b} consider this way to more seriously evaluate the model predictability on unlabeled instances. However, this setting is not a standard practice in multi-label classification, nor is it supported by LibMultiLabel. Thus we modify the scripts in LibMultiLabel to have the same evaluation setting as~\citet{IC22b}.

\section{Additional Details of Linear Methods}\label{asec:linear_methods}

The binary linear SVM is in the following form.
\begin{equation}\label{eq:binary}
\min_{\boldsymbol{w}} \frac{1}{2} \boldsymbol{w}^\top \boldsymbol{w} + C \sum_i \xi (y_i \boldsymbol{w}^\top \boldsymbol{x_i}),
\end{equation}
where $(\boldsymbol{x_i}, y_i)$ are data-label pairs in the data set, $y_i=\pm1$, $\boldsymbol{w}$ is the parameters of the linear model, and $\xi(\cdot)$ is the loss function.
The decision value function is $f(\boldsymbol{x}) = \boldsymbol{w}^\top\boldsymbol{x}$.

For one-vs-rest, please see descriptions in Section~\ref{ssec:linear_investigate}.
We follow the default setting in LibMultiLabel by using $C=1$. For more details about thresholding and cost-sensitive, please refer to the explanations in~\citet{LCL22a}.

\begin{table*}[!tb]
\centering
\begin{tabular}{l|c|c|c|c|c|c}
Property & {\sf ECtHR (A)} & {\sf ECtHR (B)} & {\sf SCOTUS} & {\sf EUR-LEX} & {\sf LEDGAR} & {\sf UNFAIR-ToS} \\
\hline
\# labels & 10 & 10 & 13 & 100 & 100 & 8 \\
$\overline{W}$ & 1,662.08 & 1,662.08 & 6,859.87 & 1,203.92 & 112.98 & 32.70 \\
\hline
\# features & 92,402 & 92,402 & 126,406 & 147,465 & 19,997 & 6,291 \\
\end{tabular}
\caption{Data statistics for LexGLUE, the benchmark considered in~\citet{IC22b}. $\overline{W}$ means the average \# words per instance of the whole set. The \# features indicates the \# TF-IDF features used by linear methods.}\label{table:data}
\end{table*}

\begin{table*}[!tb]
\centering
\begin{tabular}{l|cccc|c}
\multirow{4}{*}{Parameter} & \multicolumn{4}{c|}{LibMultiLabel} & \multirow{4}{*}{\citet{IC22b}} \\
& \multirow{3}{*}{default} & \multirow{3}{*}{tuned} & \multicolumn{2}{c|}{reproduced} \\
& & & {\sf SCOTUS} & other \\
& & & {\sf LEDGAR} & problems \\
\hline
maximum \#epochs & 15 & 15 & 20 & 15 & 20 \\
weight\_decay & 0.001 & 0 & 0 & 0 & 0 \\
patience & 5 & 5 & 5 & 5 & 3 \\
val\_metric & Micro-F1 & Micro-F1 & Micro-F1 & Micro-F1 & Micro-F1 \\
early\_stopping\_metric & Micro-F1 & Micro-F1 & loss & Micro-F1 & loss \\
learning\_rate & 5e-5 & \multirow{2}{*}{See Table~\ref{table:bert_param}} & 3e-5 & 3e-5 & 3e-5 \\
dropout & 0.1 & & 0.1 & 0.1 & 0.1 \\
\end{tabular}
\caption{Parameter differences of BERT between LibMultiLabel and~\citet{IC22b}. For the meaning of each parameter, please refer to the software LibMultiLabel.}
\label{table:bert_diff}
\end{table*}

\begin{table*}[!tb]
\centering
\tabcolsep = 0.10cm
\begin{tabular}{@{}ll|cccccc@{}}
\multicolumn{2}{@{}l|}{Parameter} & {\sf ECtHR (A)} & {\sf ECtHR (B)} & {\sf SCOTUS} & {\sf EUR-LEX} & {\sf LEDGAR} & {\sf UNFAIR-ToS} \\
\hline
\multirow{2}{*}{max\_seq\_length} & space & \multicolumn{6}{c}{[128, 512]} \\
& selected & 512 & 512 & 512 & 512 & 512 & 512 \\
\hline
\multirow{2}{*}{learning\_rate} & space & \multicolumn{6}{c}{[2e-5, 3e-5, 5e-5]} \\
& selected & 2e-5 & 3e-5 & 2e-5 & 5e-5 & 2e-5 & 3e-5 \\
\hline
\multirow{2}{*}{dropout} & space & \multicolumn{6}{c}{[0.1, 0.2]} \\
& selected & 0.1 & 0.2 & 0.1 & 0.1 & 0.2 & 0.1 \\
\end{tabular}
\caption{Hyper-parameter search space and the selected values of LibMultiLabel's tuned setting.}
\label{table:bert_param}
\end{table*}

\section{Differences Between Our Implementation of Linear Methods and~\citet{IC22b}}\label{asec:linear_details}

We summarize the implementation differences between LibMultiLabel and~\citet{IC22b} in Table~\ref{table:linear_diff}.

For the data-preprocessing part, both use scikit-learn for TF-IDF feature generations. The meanings of each parameter are listed as follows.
\begin{description}[leftmargin=0pt]
    \item[stop\_words:]
    Specify the list of stop words to be removed. For example,~\citet{IC22b} set stop\_words to ``english,'' so tokens that include in the ``english'' list are filtered.
    \item[ngram\_range:]
    Specify the range of n-grams to be extracted. For example, LibMultiLabel only uses uni-gram, while~\citet{IC22b} set ngram\_range to (1, 3), so uni-gram, bi-gram, and tri-gram are extracted into the vocabulary list for a richer representation of the document.
    \item[min\_df:]
    The parameter is used for removing infrequent tokens.~\citet{IC22b} remove tokens that appear in less than five documents, while LibMultiLabel does not remove any tokens.
    \item[max\_features:]
    The parameter decides the number of features to use by term frequency. For example,~\citet{IC22b} consider the top 10,000, 20,000, and 40,000 frequent terms as the search space of the parameter.
\end{description}
For more detailed explanations, please refer to the~\href{https://scikit-learn.org/stable/modules/generated/sklearn.feature_extraction.text.CountVectorizer.html}{TfidfVectorizer} function in scikit-learn.

The binary classification problem in~\eqref{eq:binary} is referred to as the primal form.
The optimization problem can be transferred to the dual form and the optimal solutions of the two forms lead to the same decision function.
Thus we can choose to solve the primal or the dual problem; see Table~\ref{table:linear_diff}.
For the model training, they both use the solver provided by LIBLINEAR~\citep{REF08a}.

\section{Additional Details about Model Size}\label{asec:size}

We calculate the model size of linear SVM by multiplying the number of TF-IDF features by the number of labels; see details in Table~\ref{table:data}. For BERT, we directly copy the number of parameters from~\citet{IC22b}.

\begin{table*}[!tb]
\centering
\tabcolsep = 0.20cm
\begin{tabular}{@{}l|cc|cc|cc|cc|cc|cc@{}}
\multirow{2}{*}{Method} & \multicolumn{2}{c|}{{\sf ECtHR (A)}} & \multicolumn{2}{c|}{{\sf ECtHR (B)}} & \multicolumn{2}{c|}{{\sf SCOTUS}} & \multicolumn{2}{c|}{{\sf EUR-LEX}} & \multicolumn{2}{c|}{{\sf LEDGAR}} & \multicolumn{2}{c}{{\sf UNFAIR-ToS}} \\
& \micro_f1 & \macro_f1 & \micro_f1 & \macro_f1 & \micro_f1 & \macro_f1 & \micro_f1 & \macro_f1 & \micro_f1 & \macro_f1 & \micro_f1 & \macro_f1 \\
\hline
\multicolumn{13}{@{}l}{BERT in LibMultiLabel} \\
\hline
default & 60.5 & 53.4 & 68.9 & 60.8 & 66.3 & 54.8 & 70.8 & 55.3 & 85.2 & 77.9 & 95.2 & 78.2 \\
tuned & 61.9 & 55.6 & 69.8 & 60.5 & 67.1 & 55.9 & 70.8 & 55.3 & 87.0 & 80.7 & 95.4 & 80.3 \\
{\bf reproduced} & 70.2 & 63.7 & 78.8 & 73.1 & 70.8 & 62.6 & 71.6 & 56.1 & 88.1 & 82.6 & 95.3 & 80.6 \\
\hline
\multicolumn{13}{@{}l}{BERT in~\citet{IC22b}} \\
\hline
paper & 71.2 & 63.6 & 79.7 & 73.4 & 68.3 & 58.3 & 71.4 & 57.2 & 87.6 & 81.8 & 95.6 & 81.3 \\
{\bf reproduced} & 70.8 & 64.8 & 78.7 & 72.5 & 70.9 & 61.9 & 71.7 & 57.9 & 87.7 & 82.1 & 95.6 & 80.3 \\
\end{tabular}
\caption{Micro-F1 (\micro_f1) and Macro-F1 scores (\macro_f1) for our investigation on BERT.}
\label{table:res_full}
\end{table*}

\begin{table*}[!tb]
\centering
\begin{tabular}{@{}l|c|c|c|c|c|c@{}}
Method & {\sf ECtHR (A)} & {\sf ECtHR (B)} & {\sf SCOTUS} & {\sf EUR-LEX} & {\sf LEDGAR} & {\sf UNFAIR-ToS} \\
\hline
\multicolumn{7}{@{}l}{BERT in LibMultiLabel} \\
\hline
default & 59m 48s & 1h 2m & 39m 49s & 6h 38m & 8h 44m & 47m 48s \\
tuned & 5h 8m & 5h 51m & 3h 21m & 38h 14m & 43h 48m & 4h 5m \\
{\bf reproduced} & 10h 27m & 9h 41m & 9h 26m & 6h 37m & 5h 49m & 15m 9s \\
\hline
\multicolumn{7}{@{}l}{BERT in~\citet{IC22b}} \\
\hline
paper & 3h 42m & 3h 9m & 1h 24m & 3h 36m & 6h 9m & \na \\
{\bf reproduced} & 7h 56m & 6h 59m & 7h 5m & 4h 30m & 5h 11m & 7m 3s \\
\end{tabular}
\caption{Training time for our multiple settings on BERT. The average time of running five seeds is reported.}
\label{table:time_full}
\end{table*}

\section{Additional Details about BERT Design in~\citet{IC22b}}\label{asec:bert_design}

\subsection{Standard BERT for Classification}\label{assec:stan_bert}
The setting considers the original implementation in~\citet{JD19a}. They truncate the documents to have at most 512 tokens. We then take a pre-trained BERT appended with an additional linear layer for fine-tuning.

\subsection{Document Lengths}
In Table~\ref{table:data}, we present the document length for each data set in LexGLUE, the benchmark considered in~\citet{IC22b}. For {\sf ECtHR (A)}, {\sf ECtHR (B)}, {\sf SCOTUS}, and {\sf EUR-LEX}, the document lengths all exceed 512, the length limitation of BERT. Note that the numbers are underestimated because BERT uses a sub-word tokenizer that further tokenizes some words into sub-words.

\subsection{Hierarchical BERT}\label{assec:hier_bert}
\citet{IC22b} design a variant of the standard BERT for {\sf ECtHR (A)}, {\sf ECtHR (B)}, and {\sf SCOTUS} to deal with long document lengths. The detailed steps are as follows.
\begin{itemize}
    \item Each document is split into 64 segments, where each segment contains at most 128 tokens.
    \item Each segment is then fed into BERT.
    \item The [CLS] tokens generated from each segment are collected and fed into an upper-level transformer encoder.
    \item Max pooling is applied to the output of the transformer encoder.
    \item The pooled results are then fed into a linear layer for the final prediction.
\end{itemize}

\section{Differences between the Two BERT Implementations}\label{asec:bert_methods}

We summarize the implementation differences of BERT between LibMultiLabel and~\citet{IC22b} in Table~\ref{table:bert_diff}. Here we also try to reproduce results in~\citet{IC22b} by using LibMultiLabel.

For LibMultiLabel, we explain our choices of hyper-parameters as follows.
\begin{description}[leftmargin=0pt]
    \item[default:]
    This method references the parameters chosen in an example configuration\footnote{\url{https://github.com/ASUS-AICS/LibMultiLabel/blob/master/example\_config/EUR-Lex-57k/bert.yml}} from LibMultiLabel.
    \item[tuned:]
    This method performs a parameter search and is marked as ``our BERT'' in the main paper; see Table~\ref{table:bert_param} for the search space and the chosen values.
    \item[reproduced:]
    This method aims to reproduce the BERT results from~\citet{IC22b} using LibMultiLabel. We begin with imposing the same weight\_decay, learning\_rate, and dropout values as~\citet{IC22b} and also the same validation metric. However, for other parameters, which may less affect the results, we use the same values as {\bf default} and {\bf tuned}; see Table~\ref{table:bert_diff}. Except {\sf SCOTUS} and {\sf LEDGAR}, we were able to generate similar results to those in~\citet{IC22b}. To fully reproduce the results on {\sf SCOTUS} and {\sf LEDGAR}, we try to follow every setting did in~\citet{IC22b}. Specifically, we replace the PyTorch trainer originally used in LibMultiLabel with the Hugging Face trainer adopted in~\citet{IC22b} and align some of the parameters with the ones used in~\citet{IC22b}; see a column in Table~\ref{table:bert_diff} for these two sets.
\end{description}

LibMultiLabel supports standard BERT discussed in Appendix~\ref{assec:stan_bert}. For the ``default'' and ``tuned'' settings, we directly run standard BERT. For the ``reproduced'' method, we follow~\citet{IC22b} to use hierarchical BERT explained in Appendix~\ref{assec:hier_bert} for {\sf ECtHR (A)}, {\sf ECtHR (B)}, and {\sf SCOTUS} and use standard BERT for other data sets.

\begin{table*}[!tb]
\centering
\tabcolsep = 0.20cm
\begin{tabular}{@{}l|cc|cc|cc|cc|cc|cc@{}}
\multirow{2}{*}{Method} & \multicolumn{2}{c|}{{\sf ECtHR (A)}} & \multicolumn{2}{c|}{{\sf ECtHR (B)}} & \multicolumn{2}{c|}{{\sf SCOTUS}} & \multicolumn{2}{c|}{{\sf EUR-LEX}} & \multicolumn{2}{c|}{{\sf LEDGAR}} & \multicolumn{2}{c}{{\sf UNFAIR-ToS}} \\
& \micro_f1 & \macro_f1 & \micro_f1 & \macro_f1 & \micro_f1 & \macro_f1 & \micro_f1 & \macro_f1 & \micro_f1 & \macro_f1 & \micro_f1 & \macro_f1 \\
\hline
\multicolumn{13}{@{}l}{BERT in LibMultiLabel} \\
\hline
default & 60.5 & 53.4 & 68.9 & 60.8 & 66.3 & 54.8 & 70.8 & 55.3 & 85.2 & 77.9 & 95.2 & 78.2 \\
tuned & 61.9 & 55.6 & 69.8 & 60.5 & 67.1 & 55.9 & 70.8 & 55.3 & 87.0 & 80.7 & 95.4 & 80.3 \\
\hline
\multicolumn{13}{@{}l}{BERT in LibMultiLabel (re-trained)} \\
\hline
default & 63.0 & 56.1 & 69.6 & 62.8 & 69.5 & 58.8 & 75.6 & 59.2 & 85.3 & 78.4 & 94.0 & 65.4 \\
tuned & 62.4 & 55.9 & 70.3 & 62.3 & 71.4 & 61.9 & 75.6 & 59.2 & 87.2 & 81.5 & 95.2 & 79.8 \\
\end{tabular}
\caption{A performance comparison between the setting without and with re-training.}
\label{table:res_retrain}
\end{table*}

\section{Detailed BERT Results}\label{asec:bert_res}

In Tables~\ref{table:res_full} and~\ref{table:time_full}, we respectively present the test performance and the training time.
For settings of running LibMultiLabel, see Appendix~\ref{asec:bert_methods}.
For BERT in~\citet{IC22b}, we present the following two results.
\begin{description}[leftmargin=0pt]
    \item[paper:]
    Results in the paper by~\citet{IC22b} are directly copied.
    \item[reproduced:]
    Results from our running of their scripts.\footnote{\url{https://github.com/coastalcph/lex-glue}}
\end{description}
For {\sf ECtHR (A)}, {\sf ECtHR (B)}, and {\sf SCOTUS}, because there exist some issues when running the fp16 setting in our environment, we run the code of~\citet{IC22b} by using fp32 instead.
This change causes the time difference between the ``paper'' and ``reproduced'' settings in Table~\ref{table:time_full}.

Except numbers borrowed from~\citet{IC22b}, we run five seeds for all BERT experiments and report the mean test performance over all seeds.~\citet{IC22b} also run five seeds, but their test scores are based on the top three seeds with the best Macro-F1 on validation data.

For the ``tuned'' setting, because the version of LibMultiLabel that we used does not store the checkpoint after hyper-parameter search, we must conduct the training again using the best hyper-parameters.
Thus, the total time includes hyper-parameter search and the additional training.\footnote{We run five seeds in the part of additional training. Thus, we obtain five values of the total time and report the average.}

In Appendix~\ref{asec:20news}, we give an additional case study to assess the performance of the hierarchical BERT when documents are long.

\section{Issue of Using Training, Validation, and Test Sets}\label{asec:retrain}

For each problem in LexGLUE, training, validation, and test sets are available. In our experiments, of course the test set is independent from the training process. However, some issues occur in the use of the training and validation sets.

For linear methods, in contrast to deep learning methods, they do not need a validation set for the termination of the optimization process or for selecting the iteration that yields the best model. Further, they may internally conduct cross-validation to select hyper-parameters (e.g., thresholds in the thresholding method). Therefore, we combine training and validation subsets as the new training set used by the linear methods. This is the standard setting in traditional supervised learning.

For BERT training, the validation set is used for selecting the best epoch and/or the best hyper-parameters. We follow the common practice to deploy the model achieving the best validation performance for prediction. However, in linear methods, the model used for prediction, regardless of whether internal cross-validation is needed, is always obtained by training on all available data (i.e., the combination of training and validation sets). Therefore, for BERT we may also want to incorporate the validation set for the final model training. We refer to such a setting as the re-training process. Unfortunately, an obstacle is that the optimization process cannot rely on a validation set for terminating the process or selecting the best model in all iterations. Following~\citet{IJG16a}, we consider the following setting to train the combined set.
\begin{enumerate}
    \item Record the number of training steps that leads to the best validation Micro-F1 as $e^\ast$.
    \item Re-train the final model using the combination of training and validation sets for $e^\ast$ epochs.
\end{enumerate}
BERT results without/with re-training are shown in Table~\ref{table:res_retrain}. In general, the re-training process improves the performance, especially for the data sets {\sf SCOTUS} and {\sf EUR-LEX}. However, results are slightly worse in both the default and tuned settings for the data set {\sf UNFAIR-ToS}. Thus the outcome of re-training may be data-dependent.

A comparison between linear methods and BERT with re-training shows that conclusions made earlier remain the same. Because re-training is not conducted in~\citet{IC22b}, in the main paper we report the results without re-training.

\section{A Case Study of BERT on 20 Newsgroups}\label{asec:20news}

\begin{table}[!tb]
\centering
\begin{tabular}{c|c}
Property & Value \\
\hline
\# training instances & 10,182 \\
\# validation instances & 1,132 \\
\# test instances & 7,532 \\
\# classes & 20 \\
\hline
$\overline{W}$ & 283.66 \\
$W_{\max}$ & 11,821 \\
\hline
$\overline{T}$ & 552.82 \\
$T_{\max}$ & 138,679 \\
\hline
\# documents & \multirow{2}{*}{4,927 (26.14\%)} \\
exceeding 512 tokens \\
\end{tabular}
\caption{Data statistics for 20 Newsgroups.
We conduct a 90/10 split to obtain the validation data.
$\overline{W}$/$\overline{T}$ means the average \# words/tokens per instance of the whole set, and $W_{\max}$/$T_{\max}$ means the maximum \# words/tokens of the whole set.}
\label{table:data_20news}
\end{table}

\begin{table}[!tb]
\centering
\begin{tabular}{l|cc}
Method & \micro_f1 & \macro_f1 \\
\hline\hline
\multicolumn{3}{@{}l}{Linear} \\
\hline
one-vs-rest & 85.3 & 84.6 \\
thresholding & 85.3 & 84.6 \\
cost-sensitive & 85.2 & 84.5 \\
\hline
\multicolumn{3}{@{}l}{BERT} \\
\hline
default & 84.0 & 83.3 \\
tuned & \textbf{85.6} & \textbf{84.9} \\
hierarchical & 84.9 & 84.2 \\
\end{tabular}
\caption{Experimental results of 20 Newsgroups by linear methods and BERT.
For the default setting, we follow the default parameters in Table~\ref{table:bert_diff}.
For the tuned and hierarchical setting, we use the same parameter search range as the one in Table~\ref{table:bert_param}.
Further, to process the set for the hierarchical setting, each document is split into 40 segments based on the presence of consecutive newline characters, where each segment contains at most 128 tokens.}
\label{table:res_20news}
\end{table}

\citet{YW23a} applied BERT for training the data set 20 Newsgroups~\citep{KL95a} but did not check the document length.
To assess the importance of the document length, we downloaded the 20 Newsgroups set from scikit-learn\footnote{See~\url{https://scikit-learn.org/stable/modules/generated/sklearn.datasets.fetch_20newsgroups.html} for more details. We have checked that the set used in scikit-learn is the same as the ``20 Newsgroups sorted by date'' set from the original source at~\url{http://qwone.com/~jason/20Newsgroups/}.} with default parameters.
Further, we checked the document length from the word and token levels where the tokens are obtained by the ``bert-base-uncased'' tokenizer.
The data statistics are presented in Table~\ref{table:data_20news}.
We found that the 20 Newsgroups data set includes a considerable number of documents that exceed 512 tokens.
This may be an issue because BERT can only process up to 512 tokens without further design; see Appendix~\ref{asec:bert_design} for more details.
To investigate this problem, we conducted experiments using both linear classifiers and BERT.
Results are in Table~\ref{table:res_20news}.
The observations are summarized as follows.

\begin{itemize}
	\item The results of linear classifiers do not improve by using thresholding and cost-sensitive techniques to handle class imbalance.
	The reason is that the data set has a small number of labels and a more balanced class distribution.
	In addition, linear methods are still competitive with BERT.
    \item The tuned setting of BERT has the best Micro-F1 among all the methods.
    Thus, for running BERT on this set, parameter selection seems to be important.
    Interestingly, when we considered the document length using the hierarchical methods in Appendix~\ref{assec:hier_bert}, the performance was not better than the tuned setting.
\end{itemize}

In conclusion, linear methods are still a simple and efficient solution to this problem.
For BERT, we showed that using the hierarchical setting to handle long document length may not always lead to the best performance.
The result of applying hierarchical BERT may be data-dependent.
Thus a general setting for handling long documents still need to be investigated.

\end{document}